\pdfoutput=1

\documentclass[11pt]{article}

\usepackage{ACL2023}
\usepackage{booktabs}
\usepackage{times}
\usepackage{latexsym}
\usepackage{stfloats}

\usepackage[T1]{fontenc}

\usepackage[utf8]{inputenc}

\usepackage{microtype}

\usepackage{inconsolata}

\usepackage{fancyhdr}
\usepackage[T1]{fontenc}
\usepackage[normalem]{ulem}
\usepackage{adjustbox}
\usepackage{booktabs}
\usepackage{multirow}

\usepackage[labelfont=bf]{caption}
\useunder{\uline}{\ul}{}

\newcommand{\metrics}{\textsc{Lynx}}
\newcommand{\datasets}{HaluBench}

\renewcommand{\footnotesize}{\fontsize{8pt}{10pt}\selectfont}

\title{Lynx: An Open Source Hallucination Evaluation Model}

\author{Selvan Sunitha Ravi$^{1}$, Bartosz Mielczarek$^{1}$, Anand Kannappan$^{1}$, Douwe Kiela$^{2,3}$, Rebecca Qian$^{1}$  \\ \\
{$^1$ Patronus AI $^2$ Contextual AI $^3$ Stanford University}
}

\begin{document}
\maketitle
\begin{abstract}
Retrieval Augmented Generation (RAG) techniques aim to mitigate hallucinations in Large Language Models (LLMs). However, LLMs can still produce information that is unsupported or contradictory to the retrieved contexts. We introduce \metrics, a SOTA hallucination detection LLM that is capable of advanced reasoning on challenging real-world hallucination scenarios. To evaluate \metrics, we present \datasets, a comprehensive hallucination evaluation benchmark, consisting of 15k samples sourced from various real-world domains. Our experiment results show that \metrics~outperforms GPT-4o, Claude-3-Sonnet and closed and open-source LLM-as-a-judge models on \datasets. We release \metrics, \datasets~and our evaluation code for public access.
\end{abstract}

\section{Introduction}
Large Language Models (LLMs) learn in-depth knowledge from their pre-training data \citep{petroni2019language} making them useful for knowledge-intensive downstream tasks such as Question Answering. However, their knowledge cannot be easily expanded and they often struggle with ``hallucinations'' \citep{roller2020recipes, dziri-etal-2022-origin, cao-etal-2022-hallucinated}. This has led to the adaptation of Retrieval Augmented Generation (RAG) \citep{lewis2020retrieval} systems, which enables flexibility and extensibility of LLMs to internal data stores. However, these systems are still prone to generating text that is inconsistent with the provided knowledge source \citep{mallen2022not}.

In an ideal RAG system, LLMs exhibit ``faithfulness'' by producing outputs that are grounded in the retrieved contexts. Detecting whether generated answers are faithful to the provided context is therefore critical to the success of RAG systems in production. RAGAS \citep{es2023ragas} use LLMs to generate statements from a question-answer pair and compute a faithfulness score based on how many statements are supported by the given context. Other methods involve using LLM-as-a-Judge \citep{zheng2023judging} or fine-tuning lightweight LLM judges \citep{saadfalcon2024ares} to evaluate hallucinations.

\begin{figure*}[!t]
    \centering
    \includegraphics[width=1.0\textwidth]{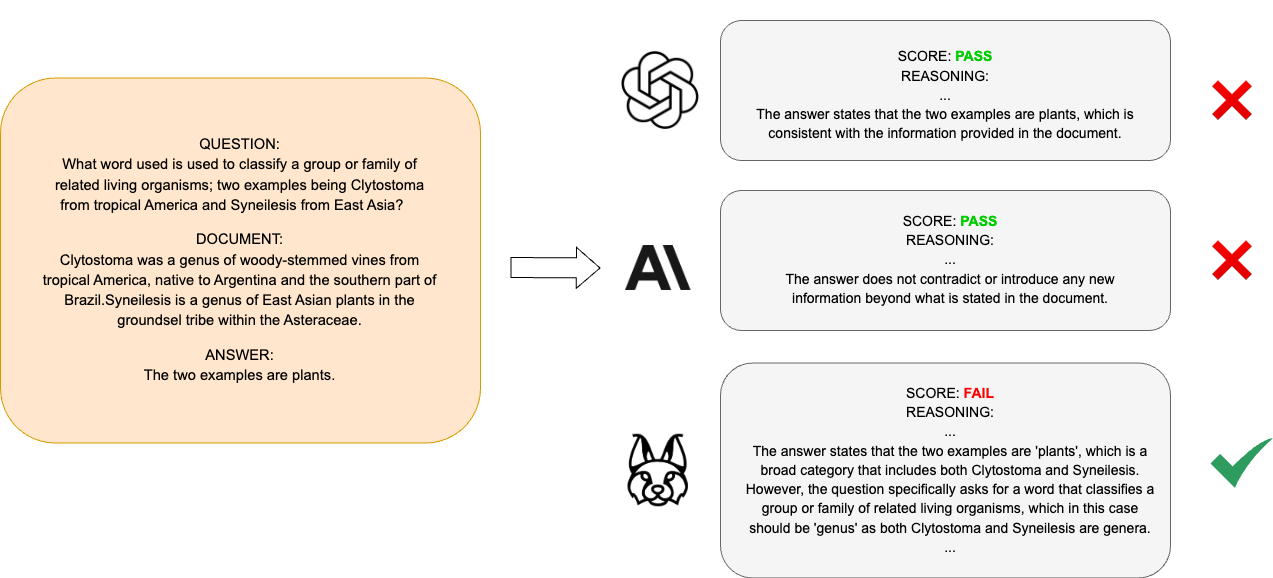}
    \caption{\looseness-1 LLM-as-a-judge responses of GPT-4o, Claude-3-Sonnet and \metrics~(70B) for a Question Answering example from HaluEval.}
    \label{fig:halueval_example}
\end{figure*}

While LLMs as judges have shown promise in automated evaluation on certain tasks \citep{zheng2024judging, zhu2023judgelm, kim2023prometheus}, hallucination detection presents a complex challenge as it requires language models to have ability to perform nuanced reasoning and disambiguation. Figure \ref{fig:halueval_example} illustrates such an example where various language models evaluate whether a Context-Question-Answer triplet contains hallucinations. GPT-4o and Claude-3-Sonnet both fail to identify that the answer, though it makes a correct statement, is not properly contextualized by the document and question.

Additionally, closed source LLMs as judges lack transparency and accessibility. Open-source LLMs still exhibit a significant gap in performance compared to closed source alternatives \citep{li2023halueval}. The gap in baseline performance between closed and open-source models increases when applied to specialized domains such as finance and medicine \citep{islam2023financebench, wu2023-nephrology}.

To address these issues, we propose \metrics~(70B) that outperforms GPT-4o and closed source LLMs in hallucination detection tasks, while being fully reproducible and open source. \metrics~(8B) produces high quality evaluations at a fraction of the size and cost of closed source LLMs. \metrics~is the first open source hallucination detection model that outperforms GPT-4o and closed source LLMs-as-Judge.

To train \metrics, we finetune Llama-3-70B-Instruct on data from multiple domains, focusing on hard-to-detect hallucinations. We source examples from existing Question Answer (QA) datasets such as CovidQA \citep{moller2020covid}, PubmedQA \citep{jin2019pubmedqa}, DROP \citep{dua2019drop} and FinanceBench \citep{islam2023financebench}, introducing perturbations to generate hallucinated answers that appear plausible but are not faithful to the context. 


Evaluating the performance of different LLMs as judges in real-world hallucination detection tasks is difficult due to the lack of a comprehensive and diverse benchmark. Halueval \citep{li2023halueval} and RAGTruth \citep{wu2023ragtruth} provide a large collection of generated and human-annotated hallucinated samples but cover limited domains. To evaluate hallucination detection systems, we construct \datasets, a large scale hallucination evaluation benchmark that consists of 15k hallucinated as well as faithful responses to questions across multiple real-world domains.

Our contributions are as follows:
\begin{itemize}
    \item We present \datasets, a hallucination evaluation benchmark of 15k samples that consists of Context-Question-Answer triplets annotated for whether the examples contain hallucinations. Compared to prior datasets, \datasets~is the first open-source benchmark containing hallucination tasks sourced from real-world domains that include finance and medicine.
    \item We train \metrics, the first open-source LLM capable of high quality, reference-free hallucination detection in RAG settings. We show that \metrics~outperforms GPT-4o, Claude-3-Sonnet and other closed and open-source models on \datasets.
    \item We propose a novel method to generate hard-to-detect hallucination examples from Question Answering tasks by applying semantic perturbations to LLM responses. We find that our perturbed examples are challenging for LLM judges, as nuanced differences in semantic meaning can lead to different reasoning outcomes.
    \item We conduct experiments to benchmark \metrics~against closed and open-source LLMs and RAG evaluation metrics. We release all models, datasets and experiment results for public access. 
    
    
\end{itemize}

\metrics\footnote[1]{HuggingFace Model: \href{https://huggingface.co/PatronusAI/Llama-3-Lynx-70B-Instruct}{https://huggingface.co/PatronusAI/Llama-3-Lynx-70B-Instruct}}and \datasets\footnote[2]{\datasets: \href{https://huggingface.co/datasets/PatronusAI/HaluBench}{https://huggingface.co/datasets/PatronusAI/HaluBench}} are available on HuggingFace. We are also releasing the training data, code and model generations on Github\footnote[3]{Github repo: \href{https://github.com/patronus-ai/Lynx-hallucination-detection}{https://github.com/patronus-ai/Lynx-hallucination-detection}}. A visualization of \datasets~is publicly available on Nomic Atlas \footnote[4]{Nomic Atlas: \href{https://atlas.nomic.ai/data/patronus-ai/halubench/map}{https://atlas.nomic.ai/data/patronus-ai/halubench/map}}

\section{Related Work}


While LLMs have shown remarkable performance in knowledge-intensive tasks such as Question Answering, one of the major drawbacks is generation of inaccurate or false information \citep{azaria2023internal, ji2023survey}. Several techniques have been proposed to evaluate hallucinations in LLMs \citep{guerreiro2022looking, lin2023llm, manakul2023selfcheckgpt} including Retrieval-Augmented Generation (RAG) \citep{lewis2020retrieval, shuster-etal-2021-retrieval-augmentation, yu-2022-retrieval, biswas-etal-2022-retrieval}. By leveraging retrieval, RAG helps LLMs acquire domain-specific knowledge and ground their outputs in factual information \citep{shuster-etal-2021-dialogue}. However, RAG systems can still suffer from hallucinations \citep{wu2023ragtruth, li2023halueval, es2023ragas, saadfalcon2024ares}. Automatic evaluation is crucial for quickly deploying these systems in new environments where creating a traditional benchmark dataset from the ground up is challenging. 

To evaluate RAG systems, LLMs have been utilized to compute metrics such as answer correctness, groundedness and the relevance of retrieved contexts \citep{zhao2019moverscore, yuan2021bartscore, liu2023-g-eval, es2023ragas, saadfalcon2024ares}. RAGAS \citep{es2023ragas} relies on a set of heuristic, hand-written prompts, while using LLMs and embedding-based similarity scores. Similarly, the EXAM \citep{sander2021exam} metric evaluates retrieval-augmented generation (RAG) systems by estimating how many exam questions a simulated QA system can correctly answer based on the generated responses. ARES constructs LLM judges using few-shot demonstrations and a bootstrapped training dataset~\citep{saadfalcon2024ares}. We find that heuristics-based metrics such as RAGAS perform poorly on hallucination tasks compared to LLM-as-judge evaluators (Table \ref{tab:results}). While ARES offers more flexibility than prior metrics, the construction of training datasets on the fly introduces significant overhead, making the approach less suitable for production settings.

A related area of research is Natural Language Inference (NLI) \citep{chen-etal-2018-neural-natural}, where the task of categorizing whether statements entail or contradict one another is similar to detecting LLM outputs that are inconsistent with provided contexts. Recent work has drawn parallels between the task of NLI and hallucination detection \citep{honovich2022true}. While most existing models are finetuned solely to output an evaluation score, \metrics~is trained using both reasoning chains and evaluation scores similar to NLI tasks, thereby improving the interpretability of the evaluation score.

Another line of research assesses the factuality of LLM responses. Several datasets have been constructed for fact extraction and verification, including FEVER \citep{fever-2021-fact} and AIS \citep{rashkin-ais}, which assesses whether outputs are attributable to identifiable sources. Prior work on automated factuality evaluation includes metrics and model based approaches \citep{ganesan2018-rouge, kryściński2019-evaluatingfactualconsistencyabstractive, sellam2020-bleurt}. More recently, \citet{wei2024-longformfactualitylargelanguage} proposed augmenting evaluator LLMs with Google Search for scoring long form factuality. Though factuality is often measured in search-based settings, TruthfulQA \citep{lin-etal-2022-truthfulqa} measures how models respond to common falsehoods and misconceptions . While factuality is important for building trust in AI systems, our work focuses on the problem of hallucination detection as it applies to RAG settings.

To evaluate the performance of models on hallucination detection, \citet{wu2023ragtruth} introduced RAGTruth, a dataset of 18k responses consisting of generations from a variety of LLMs. Similarly \citet{li2023halueval} introduced the HaluEval dataset, which contained synthetic responses generated from LLMs that were prompted to generate hallucinatory outputs. However, these datasets do not include domain specific tasks, which can be significantly more complex and more similar to real-world scenarios that users encounter in industry applications. While CRAG \citep{yang2024-crag} consists of several industry domains, the task includes external APIs for end-to-end system testing as opposed to focusing on hallucinations in provided contexts. In our construction of \datasets~we use existing datasets as well as synthetic data perturbations to construct a comprehensive hallucination evaluation benchmark that includes specialized domain specific QA tasks from finance and medicine.




\section{Methodology}
In a RAG pipeline, we first (1) Retrieve the relevant context(s) given a query, then (2) Generate an answer to the query given the retrieved context(s) with an LLM. ``Hallucinations'' \citep{jian-etal-2022-embedding, ji2023survey} occur when the generated answers are not faithful to the context (intrinsic hallucinations) or don't align with factual reality (extrinsic hallucinations). In this paper, we focus solely on intrinsic hallucination evaluation since in real-world settings, user-provided documents may contain information that conflicts with external knowledge sources. The purpose of \metrics~is to provide a reference-free metric for automated RAG evaluation, thus we consider factuality assessments out of scope for this work.

In the following sections, we describe the process for training \metrics, a SOTA hallucination detection LLM. We begin with the definition of hallucination (Section 3.1), followed by the construction process of our training and evaluation data (Section 3.2). Finally, we present experimental results on hallucination tasks sourced from real-world domains.

\subsection{Hallucination Evaluation}

For a given question $x$, we say that the LLM is hallucinating if the answer $P(x)$ is not supported by the context $C(x)$ when contextualized by the question. In practice, LLM generated answers are often inconsistent with the retrieved context \citep{li2023halueval}. In our definition of hallucination, we do not assess the relevance of the retrieved context $C(x)$ to the query $x$. If the answer, $P(x)$ is consistent with the irrelevant context, $C(x)$ we will consider the answer to be faithful to the context. Similarly, if the answer, $P(x)$ to a question, $x$ is incorrect but it states information consistent with the context, $C(x)$ it will be evaluated as faithful. 



\subsection{\datasets Construction}
We sourced examples from several existing QA datasets to build the hallucination evaluation benchmark. We constructed tuples of (\emph{question, context, answer, label}), where \emph{label} is a binary score that denotes whether the answer contains a hallucination. \datasets~consists of the following tasks:
\begin{itemize}
    \item \textbf{FinanceBench} \citep{islam2023financebench}: FinanceBench consists of 10k questions, contexts and answers over financial documents, containing both tables and bullet point lists. FinanceBench was designed to be similar to real-world financial question and answering from financial analysts. We randomly sampled 1k samples, of which 500 contain hallucinations. 
    
    \item \textbf{DROP} \citep{dua2019drop}: DROP is an English reading comprehension benchmark that assesses reasoning ability over the content of paragraphs. The dataset was crowdsourced and adversarially created. We randomly sampled 1k samples, of which 500 contain hallucinations. 
    
    \item \textbf{COVID-QA} \citep{moller2020covid}: COVID-QA consists of 2k question-answer pairs annotated by volunteer biomedical experts on scientific articles related to COVID-19. We randomly sampled 1k samples, of which 500 contain hallucinations.
    
    \item \textbf{PubMedQA} \citep{jin2019pubmedqa}: PubMedQA is a biomedical question answering (QA) dataset collected from PubMed abstracts. The task consists of answering research questions with yes/no/maybe responses. It also contains a long answer that provides evidence from the context for the response.
  
    \item \textbf{HaluEval} \citep{li2023halueval}: HaluEval is a hallucination evaluation dataset consisting of general user queries with ChatGPT responses and task-specific examples from three tasks, i.e., question answering, knowledge-grounded dialogue, and text summarization. We used the \emph{qa\_samples} subset, which contains 10k questions with knowledge from Wikipedia and question text and ground-truth answers collected from HotpotQA.
    
    \item  \textbf{RAGTruth} \citep{wu2023ragtruth}: RAGTruth is a corpus containing word-level hallucination annotations on LLM generated text. We used the test split that comprised of 900 samples, of which 160 examples contain hallucinations.
\end{itemize}

\paragraph{Construction of Hallucination Examples}
Four of the QA datasets we sourced from (DROP, FinanceBench, COVID-QA, PubMedQA) do not contain answers that are not faithful to the context. To construct unfaithful answers, we used the Context-Question-Answer to generate semantically perturbed versions of gold answers to questions. We define a semantic answer perturbation as a response that is minimally different to the gold answer, but contains an inconsistency with the context that results in a hallucination. We use GPT-4o to construct these perturbations. The prompt and generation settings are in Appendix \ref{appendix:appendix-llm}.

Let $\{q, c, x, y\}$ denote the question, context, answer and label of a given example in dataset $D$, where $y\in\{0, 1\}$. For our perturbation generator $f_p$, let $\tilde{x}\sim f_p(q, c, x)$ be the semantically altered perturbation output. Our perturbed dataset is thus

\begin{equation}
 D' = \{(q, c, \tilde{x}, 1 - y) | (q, c, x, y) \in D\}   
\end{equation}

To construct \datasets, we randomly sampled 500 examples from each of the four datasets. We then additionally sampled 500 examples and constructed a perturbed set containing hallucinations by applying the perturbation generator $f_p$. The final task consisted of a balance of positive and negative labels. See Table 2 for a breakdown of tasks in \datasets. We present some examples from \datasets~in Table \ref{tab:example_prompts}. The hallucinated answers from DROP and FinanceBench demonstrate answer perturbations. We adopted the same perturbation approach to construct training and validation datasets to finetune models for faithfulness evaluation.

\begin{table*}[t]
\vspace{-5mm}
\small
\centering
\begin{tabular}{|p{2cm}|p{13cm}|}
\toprule
  \textbf{Dataset} &  \textbf{Example}  \\
\midrule
\rule{0pt}{4pt}
HaluEval & \textbf{Context}: 750 Seventh Avenue is a 615 ft (187m) tall Class-A office skyscraper in New York City. 101 Park Avenue is a 629 ft tall skyscraper in New York City, New York. \newline \textbf{Question}: 750 7th Avenue and 101 Park Avenue, are located in which city? \newline \textbf{Answer}: 750 7th Avenue and 101 Park Avenue are located in Albany, New York. \\
\midrule
DROP & \textbf{Context}: Hoping to rebound from the road loss to the Chargers, the Rams went home for Week 9, as they fought the Kansas City Chiefs in a \"Show Me State Showdown\". The Chiefs struck first as RB Larry Johnson got a 1-yard TD run for the only score of the period.  In the second quarter, things got worse for the Rams as QB Damon Huard completed a 3-yard TD pass to TE Tony Gonzalez, while kicker Lawrence Tynes nailed a 42-yard field goal.  St. Louis got on the board with RB Steven Jackson getting a 2-yard TD run, yet Huard and Gonzalez hooked up with each other again on a 25-yard TD strike.  Rams kicker Jeff Wilkins made a 41-yard field goal to end the half.  In the third quarter, QB Marc Bulger completed a 2-yard TD pass to WR Kevin Curtis for the only score of the period, yet the only score of the fourth quarter came from Huard completing an 11-yard TD pass to TE Kris Wilson. With the loss, the Rams fell to 4-4. \newline \textbf{Question}:Which team scored the longest field goal kick of the game?   \newline \textbf{Answer}:  Rams \\
\midrule
CovidQA &  \textbf{Context}: .......An important part of CDC’s role during a public health emergency is to develop a test for the pathogen and equip state and local public health labs with testing capacity. CDC developed an rRT-PCR test to diagnose COVID-19. As of the evening of March 17, 89 state and local public health labs in 50 states...... 	    \newline    \textbf{Question}: What kind of test can diagnose COVID-19?  \newline \textbf{Answer}: rRT-PCR test      \\
\midrule
FinanceBench & \textbf{Context}:  Consolidated Statement of Income PepsiCo, Inc. and Subsidiaries Fiscal years ended December 29, 2018, December 30, 2017 and December 31, 2016 (in millions except per share amounts)  2018  2017  2016 Net Revenue \$ 64,661...... \newline  \textbf{Question}:  What is the FY2018 fixed asset turnover ratio for PepsiCo? Fixed asset turnover ratio is defined as: FY2018 revenue / (average PP\&E between FY2017 and FY2018). Round your answer to two decimal places. \newline \textbf{Answer}: 3.7\% \\
\midrule
PubmedQA & \textbf{Context}: .......The study cohort consisted of 1,797 subjects (1,091 whites and 706 blacks; age = 21-48 years) enrolled in the Bogalusa Heart Study since childhood. BP variability was depicted as s.d. of 4-8 serial measurements in childhood....... \newline \textbf{Question}: Is adult hypertension associated with blood pressure variability in childhood in blacks and whites : the bogalusa heart study? \newline \textbf{Answer}: No. Increases in BP variations as well as levels in early life are not predictive of adult hypertension, which suggests that childhood BP variability does not have a significant impact on the natural history of essential hypertension.  \\
\bottomrule
\end{tabular}
\caption{Examples of hallucinations from HaluBench. If the answer is not supported by the context, it is regarded as a hallucination.}
\label{tab:example_prompts}
\end{table*}

\paragraph{Human Annotation}
To verify that LLM generated samples in \datasets~are of high quality and that our perturber $f_p$ did actually induce hallucinations, we selected a random subset of 50 examples each from DROP, FinanceBench, CovidQA and PubMedQA for human annotation. Expert annotators manually checked the original and perturbed answers as well as reasoning provided for each example. Annotators found the data to be of relatively high quality (see Table \ref{tab:human-annotator}), with high human agreement of 0.94 across 200 samples.

We also manually annotated all examples of FinanceBench in HaluEval. We used the human annotated labels as ground truth for evaluation.

\begin{table}[h!]
\centering
\begin{tabular}{ c | c  } 
 \hline
 \textbf{Dataset} & \textbf{Score} \\ 
 \hline
 DROP  & 0.92 \\ 
 FinanceBench &  0.90\\ 
 CovidQA & 0.96\\
PubmedQA & 0.96 \\
 \hline
\end{tabular}
\caption{Agreement with human annotator for a subset of HaluBench. We use n=50 samples for each of the above datasets.}
\label{tab:human-annotator}
\end{table}




\subsection{Model Training}
\paragraph{Training Dataset Construction}
The training dataset for \metrics~consists of 2400 samples, along with 800 samples for validation. The dataset consists of demonstrations sourced from RAGTruth, DROP, CovidQA and PubMedQA. For each sub-task, we sampled 600 examples from the train split of the source dataset, of which 300 were perturbed to construct hallucinated answers. 

Chain of Thought (CoT)  has been shown to improve zero-shot performance of LLMs \citep{wei2022chain}. To distill the evaluation reasoning capabilities of GPT-4o to our finetuned open-source model, we used GPT-4o to generate reasoning for the label of each example in our training set.  We provided this as as part of the assistant response, along with the label in the instruction tuning process. The prompt to generate reasoning traces is present in Appendix \ref{appendix:appendix-llm}.

\paragraph{Self-Instruct Tuning}
We trained two models with supervised fine-tuning using the Llama-3-70B-Instruct and Llama-3-8B-Instruct checkpoints on a dataset of 2400 (question, answer, context, label) examples. Examples are formatted for instruction-tuning \citep{wei2021finetuned} in a chat-based format, where the evaluation task is provided in the instruction to the assistant, and the gold answer is the assistant response. The model is tasked to output JSON in the following format:
\begin{verbatim}
    {
        "REASONING": <reasoning provided as
                     bullet points>,
        "SCORE": <final score of PASS or FAIL>
    }
\end{verbatim}

We trained the model for 3 epochs with a learning rate of 5.0e-7 and batch size of 256. For supervised finetuning on 70B models, we trained on 32 Nvidia H100 GPUs. We used several performance optimizations including FSDP and flash attention. Our full training setup for both \metrics~(8B) and \metrics~(70B) is described in detail in Appendix \ref{appendix:training}.

\section{Results}
\subsection{Evaluation Results}
We evaluated \metrics~on \datasets~to assess its performance on hallucination detection in real world settings.

To put the results in context, we compared our proposed solution (shown as \metrics~in Table 2) with several baseline methods. We prompted the model to assess whether the response was faithful to the context, and provided the question, answer and context. We instructed the model to produce a binary score, where "FAIL" indicates that it was hallucinated and "PASS" indicates that the response was faithful. We also instructed the model to produce reasoning for the score. We used the same zero-shot prompt for all models and tasks, to ensure a fair comparison and generalization of our approach to new domains. We additionally show results for RAGAS by setting a faithfulness threshold of 50\%, where any score less than 50\% is treated as a hallucination. Results are shown in Table \ref{tab:results}. The evaluation prompt is available in Appendix \ref{appendix:appendix-llm}.


\begin{table*}[t]
\centering
\scriptsize
\begin{tabular}{l|c|c|c|c|c|c|c}
\toprule
\textbf{Model}    & \begin{tabular}[c]
{@{}l@{}}\textbf{HaluEval}\end{tabular} & \begin{tabular}[c]
{@{}l@{}}\textbf{RAGTruth}\end{tabular} & \begin{tabular}[c]
{@{}l@{}}\textbf{FinanceBench}\end{tabular} & \begin{tabular}[c]
{@{}l@{}}\textbf{DROP}\end{tabular} & \begin{tabular}[c]
{@{}l@{}}\textbf{CovidQA}\end{tabular} & \begin{tabular}[c]
{@{}l@{}}\textbf{PubMedQA}\end{tabular} & \begin{tabular}[c]{@{}l@{}}\textbf{Overall}\end{tabular}  \\
\midrule
GPT-4o  & 87.9\% & 84.3\% & \textbf{85.3\%} & 84.3\% & 95.0\% & 82.1\% & 86.5\% \\
GPT-4-Turbo  & 86.0\% & \textbf{85.0\%} & 82.2\% & 84.8\% & 90.6\% & 83.5\% & 85.0\% \\
GPT-3.5-Turbo  & 62.2\% & 50.7\% & 60.9\% & 57.2\% & 56.7\% & 62.8\% & 58.7\% \\
Claude-3-Sonnet  & 84.5\% & 79.1\% & 69.7\% & 84.3\% & 95.0\% & 82.9\% & 78.8\% \\
Claude-3-Haiku  & 68.9\% & 78.9\% & 58.4\% & 84.3\% & 95.0\% & 82.9\% & 69.0\% \\
RAGAS Faithfulness  & 70.6\% & 75.8\% & 59.5\% & 59.6\% & 75.0\% & 67.7\% & 66.9\% \\
Mistral-Instruct-7B & 78.3\% & 77.7\% & 56.3\% & 56.3\% & 71.7\% & 77.9\% & 69.4\% \\ 
Llama-3-Instruct-8B  & 83.1\% & 80.0\% & 55.0\% & 58.2\% & 75.2\%& 70.7\% & 70.4\% \\
Llama-3-Instruct-70B & 87.0\% &  \underline{83.8\%} & 72.7\% & 69.4\% & 85.0\% & 82.6\% & 80.1\% \\
\metrics~(8B) & 85.7\% & 80.0\% & 72.5\% & 77.8\% & 96.3\% & 85.2\% & 82.9\%\\  
\metrics~(70B) &  \textbf{ \underline{88.4\%}} & 80.2\% &  \underline{81.4\%} &  \underline{\textbf{86.4\%}} &  \underline{\textbf{97.5\%}} &  \underline{\textbf{90.4\%}} &  \underline{\textbf{87.4\%}} \\   
\bottomrule
\end{tabular}
\caption{Accuracy of different LLMs on \datasets. Note that DROP, CovidQA and PubMedQA contain semantically perturbed samples in addition to the original samples. The best performance among open-source models is denoted by underline and the best overall performace is denoted by bold.}
\label{tab:results}
\end{table*}

Out of all closed source and open-source models evaluated, \metrics~(70B) reports the highest accuracy on all evaluation tasks. \metrics~(70B) outperformed GPT-4o by almost one percent accuracy on average across all tasks. For domain specific tasks, this difference is even more pronounced; \metrics~(70B) is 8.3\% more accurate than GPT-4o at identifying inaccurate responses in medical answers in PubMedQA. \metrics~(8B) and \metrics~(70B) both show an increase in accuracy on all tasks compared to the baseline Llama 3 models, with the finetuned 70B model resulting in a 7.8\% increase in average accuracy. When compared to closed source LLMs, \metrics~outperforms GPT-3.5-Turbo by an even wider margin, with an average increase of 27.6\% across all tasks. \metrics~(70B) is the best performing model overall, with 87.4\% accuracy on \datasets. GPT-3.5-Turbo showed the lowest accuracy out of all models evaluated, with 58.7\% accuracy averaged across all tasks.




\section{Conclusion}
As RAG systems continue to rise in popularity, automated reference-free evaluation of RAG systems is critical for the safe deployment of RAG systems at scale. We propose an evaluation model that assesses faithfulness of model responses in reference-free settings, which has important implications in business contexts ranging from detecting erroneous responses in financial Q\&A to preventing misinformation in medical AI assistants. Our results show that \metrics~outperforms industry and academic alternatives on \datasets.

We have introduced \datasets, a comprehensive hallucination evaluation benchmark that contains annotations of the faithfulness of textual responses across several real-world domains. \datasets~is unique for containing balanced distributions of positive and negative examples, and for a high percentage of examples grounded in real-world contexts. \datasets~consists of challenging hallucination detection examples, and shows high agreement with human annotations.

Lastly, we are open sourcing our evaluation datasets and model outputs, along with human annotations. The \metrics~model is lightweight and easy to use, and can provide developers of RAG systems with useful insights, especially in the absence of ground truth annotations.

\section{Limitations and Future Work}

\paragraph{Failures outside of LLM Generation}
In real world deployments of RAG systems, there are often failures outside of RAG systems that can result in inaccuracies in LLM outputs. A common failure in RAG systems outside of LLM generation is in the retrieval component, where the retriever does not return relevant contexts to the query. This can result in downstream hallucinations, as the context provided to the generation model does not contain sufficient information to address the input.

Other sources of failures in RAG systems unrelated to the LLM include pre-processing and post-processing steps, database queries and inconsistencies in data sources. In particular, source documents that contain conflicting information, or misinformation, present a challenge for failure detection due to its ambiguity. The resolution of conflicting information in fact checking tasks is a continued area of research. We leave an in depth exploration of improving retrieval modules to future work.

\paragraph{Multilingual Coverage}
The bulk of datasets used in \datasets~is in English, which presents a limitation in real-world applications that include multilingual inputs and contexts. We hope to enhance coverage and diversity in \datasets~and training data by incorporating non-English and low resource languages in future extensions.

\paragraph{Summarization and Other NLP Tasks}
\datasets~is focused on Question Answering tasks due to the prevalence of chat-based interfaces used by knowledge workers in industry RAG applications. An area for future work is extending \metrics~to additional NLP domains, including abstractive summarization tasks where LLM produced summaries may contain inconsistent information with the source document.

\paragraph{Truthfulness and World Knowledge}
\metrics~focuses on the problem of hallucination detection. The assessment of truthfulness and factuality is also an important factor of consideration, and requires the incorporation of external knowledge sources as world knowledge.

\paragraph{Natural Language Inference}
An interesting area for future work involves applying \metrics~to NLI tasks. Since the problem of hallucination detection is closely related to NLI, a strong hallucination detection model is likely capable of performing reasoning in other NLP domains. We leave research on the relationship between evaluation tasks and other NLP tasks to future work.

\section{Acknowledgements}
We would like to thank Brandon Cui, Connor Jennings from Databricks Mosaic AI, Candace Ross from Meta AI and Andriy Mulyar from Nomic AI for their feedback and contributions to the paper. We additionally thank the Mosaic Databricks AI team for their support in the development of \metrics. We also thank Nino Scherrer for his contributions to the research process. Finally, we want to thank our partners, Nvidia and Nomic AI for their support of \metrics~and \datasets.

\bibliography{custom, anthology}
\bibliographystyle{acl_natbib}

\clearpage
\appendix

\newpage
\section{Prompts}
\label{appendix:appendix-llm}

\subsection{Data Generation}
For generating the perturbed answers, we use the following prompt with GPT-4o with temperature=0.

\begin{verbatim}
QUESTION:
{question}

GOLD_ANSWER:
{gold_answer}

EVIDENCE_TEXT:
{evidence_text}

How can we change the GOLD_ANSWER subtly
such that it would be wrong? The perturbed
answer  should still give the impression
of a  valid answer, but inspection of 
the  EVIDENCE_TEXT would reveal that the
perturbed answer is factually wrong.
Output the new answer and change made
in JSON format with the key 'new_answer'
and 'change_made'.

\end{verbatim}

To generate the reasoning chains, we use the following prompts with GPT-4o with temperature=0. \\

I. For perturbed samples: \\ 

Below is the System Prompt:
\begin{verbatim}
You are given a QUESTION, CONTEXT, 
CHANGE_MADE, GOLD_ANSWER and ANSWER.
Explain why the ANSWER is not faithful
to the CONTEXT, given the QUESTION.
CHANGE_MADE specifies the change
made to the GOLD_ANSWER which made 
the ANSWER not faithful. Do not refer
explicitly to the words 'CHANGE_MADE'
or 'GOLD_ANSWER' in your reasoning. 
Generate your reasoning in JSON format:

{"REASONING": "<your reasoning steps as
bullet points>"}
\end{verbatim}

Below is the User Prompt:
\begin{verbatim}
<QUESTION>
{question}
</QUESTION>

<CONTEXT>
{context}
</CONTEXT>

<CHANGE_MADE>
{change_made}
</CHANGE_MADE>

<GOLD_ANSWER>
{answer}
</GOLD_ANSWER>

 <ANSWER>
{new_answer}
</ANSWER>

\end{verbatim}

II. For original samples: \\

System Prompt:
\begin{verbatim}
You are given a QUESTION, CONTEXT, ANSWER.
Explain the similarities between the CONTEXT
and the ANSWER. Reason about why the
ANSWER is faithful to the CONTEXT given
the QUESTION. Generate your reasoning in
JSON format:

{"REASONING": "<your reasoning steps as
bullet points>"}

\end{verbatim}

User Prompt: 
\begin{verbatim}
<QUESTION>
{question}
</QUESTION>

<CONTEXT>
{context}
</CONTEXT>

 <ANSWER>
{answer}
</ANSWER>

\end{verbatim}

\subsection{Evaluation}

We use the following prompt for instruction fine-tuning as well as for evaluation of models:

User Prompt:
\begin{verbatim}
Given the following QUESTION, DOCUMENT 
and ANSWER you must analyze the provided
answer and determine whether it is 
faithful to the contents of the DOCUMENT.
The ANSWER must not offer new information
beyond the context provided in the DOCUMENT.
The ANSWER also must not contradict
information provided in the DOCUMENT. 
Output your final verdict by strictly
following this format: "PASS" if the
answer is faithful to the DOCUMENT
and "FAIL" if the answer is not 
faithful to the DOCUMENT. Show your
reasoning.

--
QUESTION (THIS DOES NOT COUNT
AS BACKGROUND INFORMATION):
{question}

--
DOCUMENT:
{context}

--
ANSWER:
{answer}

--

Your output should be in JSON FORMAT with
the keys "REASONING" and "SCORE":
{{"REASONING": <your reasoning as
bullet points>, "SCORE": <your final score>}}

\end{verbatim}

\section{Training and Evaluation Details}
\label{appendix:training}
\subsection{Setup}
For \metrics~, we do mixed precision training with flash attention. We use a cosine scheduler with warmup. warmup steps is set to 100. We use lionw optimizer with $\beta_{1} = 0.9 $ and $\beta_{2} = 0.95 $ and norm gradient clipping with threshold=1.0. FSDP is used with FULL\_SHARD strategy and activation\_checkpointing enabled.

For evaluating the 70B models, we use vLLM on 8 H100s with tensor\_parallel = 8. For evaluating the 8B variants, we use model and data sharding with accelerate. We use HuggingFace pipelines for the generations, with greedy decoding and max\_new\_tokens = 600.

\subsection{Llama-2-13B-Chat Evaluation}

We observe that the Llama-2-13B-Chat model does not output JSON data or adhere to the response structure requested in the prompt. However, after finetuning the model, we are able to parse responses to extract REASONING and SCORE. The results are present in Table \ref{tab:llama2chat}.

\subsection{Training with extended datasets}

As \metrics~(70B) performed worse than Llama-3-Instruct-70B on the RAGTruth test split, we extended the training data to include 2k samples from RAGTruth. We finetuned Llama-3-Instruct-70B on this extended dataset. The results are reported in Table \ref{tab:extendedragtruth_performance}. While the performance increases on the RAGTruth split, we see a slight decrease in performance on the other splits. The overall performance gain with the extended RAGTruth dataset is $\sim 0.4\%$.

\begin{table*}[t]
\centering
\scriptsize
\begin{tabular}{l|c|c|c|c|c|c|c}
\toprule
\textbf{Model}    & \begin{tabular}[c]
{@{}l@{}}\textbf{HaluEval}\end{tabular} & \begin{tabular}[c]
{@{}l@{}}\textbf{RAGTruth}\end{tabular} & \begin{tabular}[c]
{@{}l@{}}\textbf{FinanceBench}\end{tabular} & \begin{tabular}[c]
{@{}l@{}}\textbf{DROP}\end{tabular} & \begin{tabular}[c]
{@{}l@{}}\textbf{CovidQA}\end{tabular} & \begin{tabular}[c]
{@{}l@{}}\textbf{PubMedQA}\end{tabular} & \begin{tabular}[c]{@{}l@{}}\textbf{Overall}\end{tabular}  \\
\midrule
Llama-3-Instruct-70B & 87.0\% & 83.8\% & 72.7\% & 69.4\% & 85.0\% & 82.6\% & 80.1\% \\

Llama-3-Instruct-70B (RAGTruth+) & \textbf{88.8\%} & \textbf{85.8\%} & 81.2\% & {85.3\%} & {96.9\%} & {88.8\%} & \textbf{87.8\%} \\   

\metrics~(70B) & {88.4\%} & 80.2\% & \textbf{81.4\%} & \textbf{86.4\%} & \textbf{97.5\%} & \textbf{90.4\%} & {87.4\%} \\   

\bottomrule
\end{tabular}
\caption{Accuracy of Llama-3-Instruct-70B model when finetuned on additional RAGTruth samples (denoted by RAGTruth+).}
\label{tab:extendedragtruth_performance}
\end{table*}

\begin{table*}[t]
\centering
\scriptsize
\begin{tabular}{l|c|c|c|c|c|c|c}
\toprule
\textbf{Model}    & \begin{tabular}[c]
{@{}l@{}}\textbf{HaluEval}\end{tabular} & \begin{tabular}[c]
{@{}l@{}}\textbf{RAGTruth}\end{tabular} & \begin{tabular}[c]
{@{}l@{}}\textbf{FinanceBench}\end{tabular} & \begin{tabular}[c]
{@{}l@{}}\textbf{DROP}\end{tabular} & \begin{tabular}[c]
{@{}l@{}}\textbf{CovidQA}\end{tabular} & \begin{tabular}[c]
{@{}l@{}}\textbf{PubMedQA}\end{tabular} & \begin{tabular}[c]{@{}l@{}}\textbf{Overall}\end{tabular}  \\
\midrule
Llama-2-Chat-13B & 3.1\% & 5.1\% & 4.4\% & 2.6\% & 2.0\% & 1.1\% & 3.3\% \\

Llama-2-Chat-13B (Finetuned) & \textbf{79.3\%} & \textbf{77.6\%} & \textbf{62.9\%} & \textbf{76.4\%} & \textbf{88.8\%} & \textbf{81.8\%} & \textbf{77.8\%}\\     

\bottomrule
\end{tabular}
\caption{Performance of Llama-2-Chat-13B model on HaluBench. Finetuning improves parsability of the responses.}
\label{tab:llama2chat}
\end{table*}

\end{document}